\documentclass[conference]{IEEEtran}
\IEEEoverridecommandlockouts

\usepackage{cite}
\usepackage{amsmath,amssymb,amsfonts}
\usepackage{algorithmic}
\usepackage{graphicx}
\usepackage{textcomp}
\usepackage[table,xcdraw]{xcolor} 
\usepackage{multirow}
\usepackage{cleveref}
\usepackage[labelformat=simple]{subcaption} 
\usepackage{adjustbox}
\usepackage{subcaption}

\usepackage[T1]{fontenc}
\def\BibTeX{{\rm B\kern-.05em{\sc i\kern-.025em b}\kern-.08em
    T\kern-.1667em\lower.7ex\hbox{E}\kern-.125emX}}
\begin{document}

\newcommand{\todo}[1]{{\color{red}#1}}
\newcommand{\TODO}[1]{\textbf{\color{red}[TODO: #1]}}

\newcommand{\etal}{\textit{et al.}}

\definecolor{dark-gray}{gray}{0.20}
\newcommand{\pub}[1]{{\color{dark-gray}{\tiny{[{#1}]}}}}
\newcommand{\red}[0]{\textcolor{red}}
\newcommand{\blue}[0]{\textcolor{blue}}
\newcommand{\gray}[0]{\textcolor{gray}}
\newcommand{\rot}[0]{\rotatebox{90}}

\Crefname{figure}{Fig.}{Figs.}

\newtheorem{definition}{Definition} 
\newtheorem{theorem}{Theorem} 
\newtheorem{theorem_supp}{Theorem} 
\newtheorem{lemma}{Lemma} 
\newtheorem{lemma_supp}{Lemma} 
\newtheorem{proof}{Proof}

\definecolor{1st}{HTML}{acdbdf}
\definecolor{2nd}{HTML}{f0ece2}

\title{Semantic Data Augmentation Enhanced \\ Invariant Risk Minimization  for   Medical Image Domain Generalization
    \thanks{$^*$ Co-corresponding authors. This work was supported by the Chengdu Science and Technology Bureau under Grant 2022-YF04-00078-JH.}
}

\author{
    \IEEEauthorblockN{Yaoyao Zhu$^{a,b}$, Xiuding Cai$^{a,b}$, Yingkai Wang$^{a,b}$, Yu Yao$^{a,b}$, Xu Luo$^{a,b*}$, Zhongliang Fu$^{a,b*}$}
    \IEEEauthorblockA{$^a$  Chengdu Institute of Computer Application, Chinese Academy of Sciences, Chengdu, 610213, China}
    \IEEEauthorblockA{$^b$  University of Chinese Academy of Sciences, Beijing, 101408, China}
    \IEEEauthorblockA{\{zhuyaoyao19, caixiuding20, wangyingkai22, luoxu18\}@mails.ucas.ac.cn}
    \IEEEauthorblockA{\{casitmed2022\}@163.com, \{fzliang\}@casit.com.cn}
}
\maketitle

\begin{abstract}
    Deep learning has achieved remarkable success in medical image classification.
    However, its clinical application is often hindered by data heterogeneity caused by variations in scanner vendors, imaging protocols, and operators.
    Approaches such as invariant risk minimization (IRM) aim to address this challenge of out-of-distribution generalization.
    For instance, VIRM improves upon IRM by tackling the issue of insufficient feature support overlap, demonstrating promising potential.
    Nonetheless, these methods face limitations in medical imaging due to the scarcity of annotated data and the inefficiency of augmentation strategies.
    To address these issues, we propose a novel domain-oriented direction selector to replace the random augmentation strategy used in VIRM.
    Our method leverages inter-domain covariance to guide augmentation direction, guiding data augmentation towards the target domain.
    This approach effectively reduces domain discrepancies and enhances generalization performance.
    Experiments on a multi-center diabetic retinopathy dataset demonstrate that our method outperforms state-of-the-art approaches, particularly under limited data conditions and significant domain heterogeneity.
    The code for our method has been made publicly available at https://github.com/YaoyaoZhu19/MedVIRM.
\end{abstract}


\section{Introduction}
\label{sec:intro}

Deep learning has shown remarkable capabilities and broad applicability in medical image analysis \cite{10601163, 10374392}.
In the past decade, significant progress has been made in tasks such as organ segmentation, tumor classification, and disease progression prediction \cite{7404017}.
Despite these advancements, real-world clinical applications often require robust performance across diverse datasets from multiple institutions \cite{multi_center}.
This emphasizes the importance of addressing data heterogeneity, which arises from variations in scanner vendors, imaging protocols, and operators \cite{9837077}.
Such heterogeneity leads to substantial variations in probability distributions across clinical datasets, significantly impacting the performance of deep learning models \cite{liu2020ms, wang2020domainmix, li2021simple}.
As a result, when deep neural network models are applied to new, unseen data from different domains, their limited generalization ability restricts their practical applicability in clinical settings, where generalization is crucial \cite{liu2020green, dgdr2023}.

\begin{figure}
    \centering
    \includegraphics[width=0.45\textwidth]{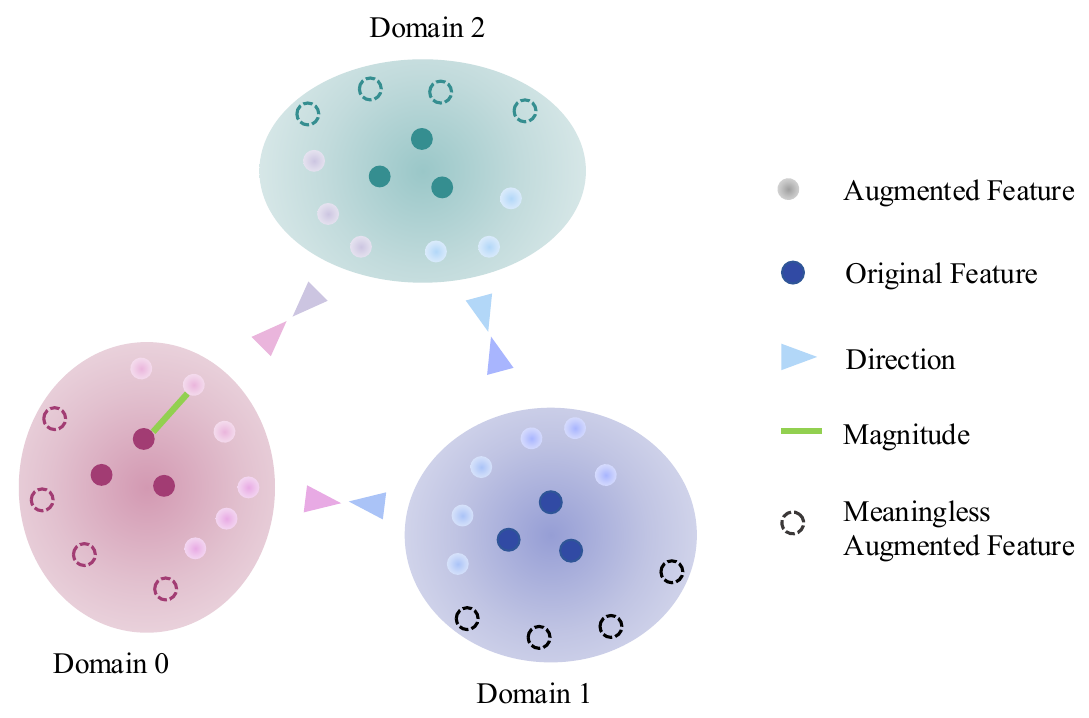}
    \caption{
        Data augmentation methods expand feature overlap.
        However, the relatively small scale of medical image datasets and the clustered nature of features often result in many meaningless directions when random data augmentation is applied.
    }
    \label{fig:meaningless}
\end{figure}

Domain generalization has recently attracted significant attention as a potential solution to this challenge \cite{wang2022generalizing, zhou2022domain}.
Existing approaches can be broadly categorized into data-centric, feature-centric, and model-centric methods.
Data-centric methods focus on data augmentation to enhance domain diversity, including techniques such as Mixup \cite{zhang2017mixup} and style augmentation-based generation \cite{li2023intra}, which improve model robustness by simulating unseen scenarios \cite{concutmix}.
Feature-centric methods, such as invariant risk minimization (IRM) \cite{arjovsky2019invariant}, aim to learn domain-invariant features to improve generalization across different domains \cite{chen2022pair, lin2022birm, li2022invariant}.
However, while vicinal invariant risk minimization (VIRM) \cite{zhu2024enlarging} shows promise in large-scale datasets,
its random direction selection strategy often results in inefficient or meaningless augmentations (as shown in \Cref{fig:meaningless}) in the context of medical imaging, where annotated data is scarce and expensive.
Random direction selection may fail in medical image analysis, as augmentations without explicit alignment toward the target domain fail to bridge domain gaps.

This issue is primarily because medical image datasets are usually smaller in scale compared to natural images and exhibit high variability.
Although other domain generalization algorithms tailored for medical image analysis, such as GDRNet \cite{dgdr2023}, have been proposed, their main limitation lies in their strong specialization.
This specialization makes them less adaptable to the diverse imaging modalities encountered in medical images, thereby restricting their ability to generalize across different types of medical imaging data.

\begin{figure*}
    \centering
    \includegraphics[width=\textwidth]{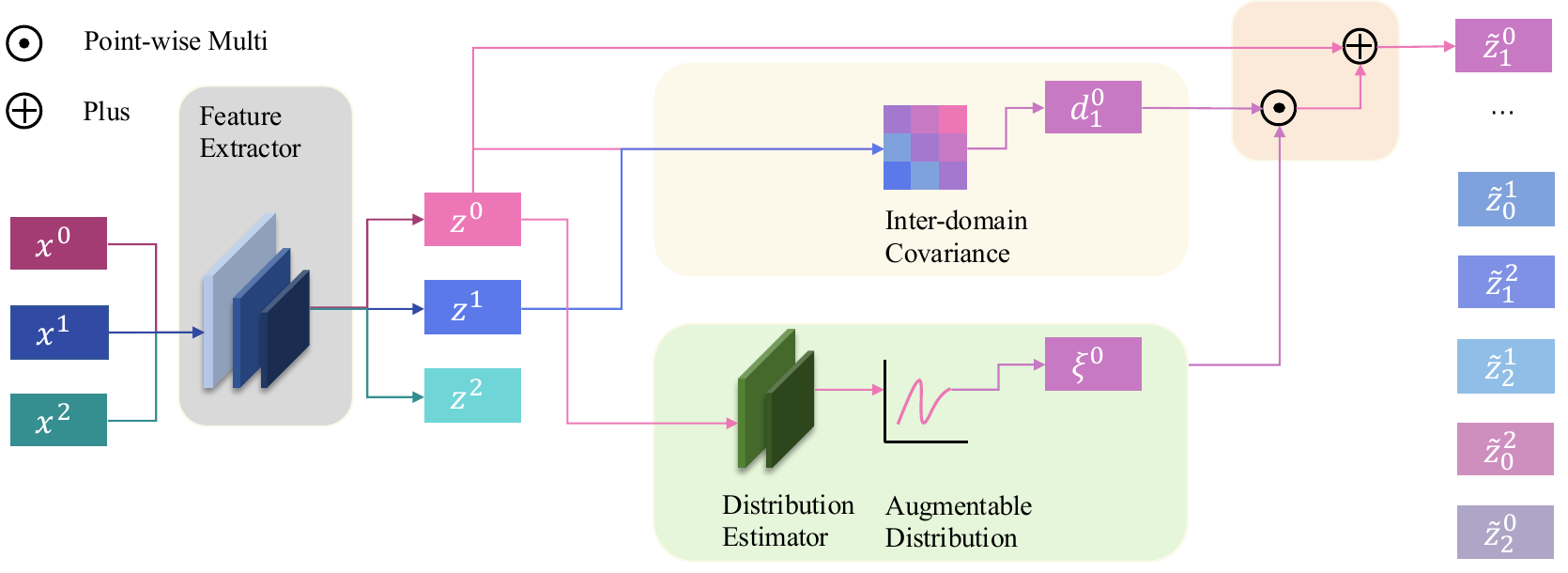}
    \caption{Overview of the proposed method framework.
        Our method includes an inter-domain covariance-guided direction selector and a distribution estimator that augmentable the range of distribution.
        $\mathbf{x}^i$ denotes the input image from domain $i$, $\mathbf{z}^i$ denotes the feature representation of $\mathbf{x}^i$,
        $\mathbf{d}^i_j$ denotes the augment direction for domain $i$ to domian $j$,
        and $\tilde{z}^i_j$ denotes the augmented feature representation of $\mathbf{z}^i$ in domain $j$.
    }
    \label{fig:framework}
\end{figure*}

To address these challenges, we propose a novel domain direction selector, guided by inter-domain covariance, to replace the random augmentation strategy used in VIRM.
Unlike VIRM, which aims to improve the overlap of representations between domains through random augmentations, our method dynamically guides each augmentation toward the target domain.
This ensures that augmented samples contribute meaningfully to reducing domain discrepancies and enhancing domain generalization.
The proposed direction selector adjusts the augmentation direction based on the inter-domain covariance, ensuring that the augmented samples effectively reduce domain discrepancies.
The overall framework of our proposed method is illustrated in \Cref{fig:framework}.
We evaluate our method on a medical image domain generalization dataset: Diabetic Retinopathy.
Experimental results demonstrate that our approach consistently outperforms state-of-the-art methods, particularly in scenarios with limited data and significant domain heterogeneity.
These findings highlight the robustness and practicality of our framework, providing a solid foundation for its broader application in medical imaging tasks.

\section{Related Work}
\label{sec:related_work}

Domain generalization (DG) has emerged as a critical area of research to develop models that generalize well across different domains, particularly in the context of medical image datasets.
A variety of techniques have been proposed in the literature to address domain shifts, which arise due to factors such as differences in imaging equipment, acquisition protocols and patient demographics.
These approaches can be broadly classified into three categories: data-centric methods, feature-centric methods, and model-centric methods.
In this section, we review these categories in the context of medical image analysis and discuss the strengths and limitations of each.

\subsection{Data-Centric Methods}

Data-centric approaches focus on enhancing domain diversity through data augmentation strategies, enabling the model to learn from a broader range of scenarios. These methods simulate unseen domain variations, which helps improve generalization performance. Popular techniques include:
Mixup: A data augmentation method that generates new samples by combining pairs of images and their corresponding labels, fostering improved model robustness \cite{zhang2017mixup}.
StyleGAN: A generative approach that synthesizes diverse data variations by learning the distribution of image styles across different domains \cite{karras2019style}.
While these methods contribute to improved model performance by simulating a variety of scenarios, they often face challenges in medical image analysis, where annotated data is limited and expensive. The limited availability of annotated data in medical domains reduces the effectiveness of random data augmentation strategies.

\subsection{Feature-Centric Methods}

Feature-centric methods aim to learn domain-invariant features that generalize well across multiple domains. These methods focus on identifying common representations shared among domains to mitigate domain shifts and improve performance on unseen data.
Invariant Risk Minimization (IRM): IRM seeks to minimize the risk of domain-specific variations by learning features that are invariant across different domains \cite{arjovsky2019invariant}.
Domain-Adversarial Neural Networks (DANN): DANN employs adversarial training to align feature distributions between domains, encouraging the network to extract domain-invariant features \cite{ganin2016domain}.
VIRM: The recent method VIRM introduces a random augmentation strategy to enhance domain overlap and improve invariant feature learning \cite{zhu2024enlarging}.
While VIRM shows promise in large-scale datasets, its random direction selection often results in inefficient augmentations for medical imaging, where annotated data is scarce and highly variable.
While feature-centric methods have demonstrated success in improving generalization across domains, they can struggle with medical imaging tasks where domain overlap is minimal or where data is highly heterogeneous.
This often leads to suboptimal feature alignment.

\subsection{Model-Centric Methods}

Model-centric methods focus on designing novel model architectures or learning strategies to handle domain shifts more effectively.
These methods typically incorporate domain-specific information into the learning process or employ meta-learning strategies to improve model adaptation.
Meta-Learning: Meta-learning approaches such as Model-Agnostic Meta-Learning (MAML) aim to train models that can quickly adapt to new domains with minimal data \cite{finn2017model}.
Domain Alignment Networks: These models introduce domain-specific components into the architecture to align features across domains while preserving domain-specific characteristics \cite{li2018domain}.
Although model-centric methods offer flexibility and adaptability, they often require large-scale datasets for training and can be computationally expensive, which limits their applicability in medical imaging tasks where labeled data is scarce.

\subsection{Domain Generalization in Medical Imaging}

Recent research has focused on applying domain generalization techniques specifically to medical image analysis.
These studies aim to address challenges such as variations in imaging modalities, acquisition protocols, and limited annotated data.
Key advancements in this area include:
GREEN\cite{liu2020green} method utilizes a Graph Convolutional Network to model class dependencies in diabetic retinopathy grading, addressing label uncertainty and improving the classification results through a residual re-ranking mechanism.
GDRNet \cite{dgdr2023} effectively reduces domain discrepancies and enhances out-of-distribution generalization, particularly under conditions of limited data and significant domain heterogeneity. However, its specificity limits its generalization to other medical imaging tasks with varying imaging principles and modalities.

\section{Method}
\label{sec:method}

\subsection{Problem Definition}
\label{sec:medvirm}

Deep learning models excel at linearizing features \cite{upchurch2017deep}, enabling them to generate semantically different features by performing linear transformations on deep features while preserving the original label \cite{ISDA}. Zhu \etal \cite{zhu2024bayesian} redefined semantic data augmentation by decomposing it into two components: semantic direction and semantic strength. The semantic direction specifies the features to be altered, enabling transformation into different semantics, while semantic strength controls the magnitude of the shift, as excessive augmentation can alter the label.
Therefore, the medical semantic data augmentation module proposed in this paper is defined as:
\begin{equation}
	\begin{aligned}
		\tilde{\mathbf{z}}_i = \mathbf{z}_i + \mathbf{d} \odot \mathbf{\xi},
	\end{aligned}
	\label{eq:sda}
\end{equation}
where $\mathbf{z}_i = \mathbf{\Phi}(\mathbf{x}_i)$, $\mathbf{d}$ is the augment direction, $\mathbf{\xi} \sim p(\mathbf{\xi}|\mathbf{z}_i)$ is the augment magnitude, and $\odot$ denotes the element-wise product.

Vicinal Invariant Risk Minimization (VIRM) \cite{zhu2024enlarging} is a domain generalization method aiming to minimize the expected loss across multiple training domains by learning domain-invariant features. VIRM introduces a Bayesian semantic data augmentation \cite{zhu2024bayesian} strategy to enhance domain overlap and improve invariant feature learning. VIRM is formulated as the following optimization problem:

\begin{definition}[VIRM]
	\label{def:virm}
	Given data representation $\mathbf{\Phi}: \mathcal{X} \rightarrow \mathcal{H}$, classifier $\mathbf{w}: \mathcal{H} \rightarrow \mathcal{Y}$, the VIRM objective minimizes the expected loss, as defined below:
	\begin{equation}
		\begin{aligned}
			 & \min_{\mathbf{\Phi, \mathbf{w}}}
			\sum_{e \in  \mathcal{E}_{\mathrm{tr}}} \sum_{\mathbf{x}_i \in e} \ell_e(\mathbf{w} (\mathbf{z}_i + \mathbf{d} \odot \mathbf{\xi}), y_i),                                                                     \\
			 & \mathrm{s.t.} \quad \mathbf{w} \in \underset{\bar{\mathbf{w}}:\mathcal{H}\to\mathcal{Y}}{\operatorname*{\arg\min}} \, \ell_e(\bar{\mathbf{w}}\circ\mathbf{\Phi}),\forall e \in \mathcal{E}_{\mathrm{tr} },
		\end{aligned}
	\end{equation}
	where $\mathbf{\xi}$ is the translation magnitude sampled from the augmentable distribution $p(\mathbf{\xi} | \mathbf{z}_i)$, and $\mathbf{d}$ is the augment direction.
\end{definition}

Thus, the proposed method includes two parts: the invariant risk and the medical semantic data augmentation (MedSDA) module.

\subsection{MedSDA Module}
\label{sec:medsda}
The MedSDA module consists of two components: the director (\Cref{sec:director}) and the estimator (\Cref{sec:estimator}) as shown in \Cref{fig:framework}. The director is responsible for selecting the augment direction $\mathbf{d}$, while the estimator estimates the augmentable distribution $p(\mathbf{\xi}|\mathbf{z}_i)$. These components are jointly trained to minimize the loss function of the MedSDA module. The MedSDA module is designed to enhance the feature overlap between domains, improving the generalization performance of the model. Therefore, $\tilde{\mathbf{z}}_i$ is generated using \Cref{eq:sda}.

\subsection{Director}
\label{sec:director}
We employ inter-domain covariance as the guiding principle for selecting augmentation directions. Let $x \in \mathcal{X}$ be the input features with domains $\mathcal{D}$, where $x$ has shape $(N, \text{out\_features})$ and $\text{domains} \in \mathcal{D}$ has shape $(N,)$. The covariance between the domains is computed as:
\begin{equation}
	C_{e} = \text{Cov}(\mathbf{x} | e), \quad \forall e \in \mathcal{E}_{\mathrm{tr} }.
\end{equation}

The covariance difference for each domain is given by:
\begin{equation}
	\Delta C_e = C_e - \mathbb{E}_{\mathcal{D}}[C], \quad \forall e \in \mathcal{E}_{\mathrm{tr} },
\end{equation}
where $\mathbb{E}_{\mathcal{D}}[C]$ represents the mean covariance across all domains.

The direction of the maximum difference for each domain is:
\begin{equation}
	\mathbf{d}_e = \arg\max_{d \in \mathcal{H}} |\Delta C_e[d]|.
\end{equation}

The binary direction selection is:
\begin{equation}
	\mathbf{d}_{e}^* =
	\begin{cases}
		1, & \text{if} \; |\Delta C_e[d]| > \mathbb{E}[\Delta C_e], \\
		0, & \text{otherwise.}
	\end{cases}
\end{equation}

\subsection{Estimator}
\label{sec:estimator}
In VIRM implementation, we estimate the augmentable distribution $p(\mathbf{\xi} |\mathbf{z}_i)$ while ensuring $\tilde{y}_i = y_i$. We introduce model $q_{\phi_e}(\mathbf{\xi}|\mathbf{\mathbf{z}}_i)$ to approximate the distribution $p(\mathbf{\xi} |\mathbf{z}_i)$ of domain $e$. The Kullback-Leibler (KL) divergence measures the similarity between these two distributions, aiming to make $q_{\phi_e}(\mathbf{\xi} |\mathbf{z}_i)$ closely match $p(\mathbf{\xi} |\mathbf{z}_i)$ by maximizing the KL divergence. Thus, our optimization goal of the SDA module is defined as:
\begin{equation}
	\label{eq:bsda}
	\phi_e = \underset{\phi_e}{\arg\max} D_{KL}(q_{\phi_e}(\mathbf{\xi}|\mathbf{\mathbf{z}}_i) || p(\mathbf{\xi}|\mathbf{\mathbf{z}}_i)).
\end{equation}

Another challenge with \Cref{eq:bsda} arises when domain-specific estimators are introduced, as they prevent feature sharing across domains, hindering the transformation of domain-invariant features. Although using shared estimators across domains can address this issue, it increases the model's optimization complexity. Thus, our domain-shared SDA module is defined as:
\begin{equation}
	\label{eq:sdsda}
	\phi = \underset{\phi}{\arg\max} D_{KL}(q_{\phi}(\mathbf{\xi}|\mathbf{\mathbf{z}}_i) || p(\mathbf{\xi}|\mathbf{\mathbf{z}}_i)).
\end{equation}

While \Cref{eq:sdsda} estimates the distribution $p(\mathbf{\xi}|\mathbf{\mathbf{z}}_i)$, additional constraints on the augmented features are essential to maintain label consistency. Specifically, consistency is constrained as:
\begin{equation}
	\label{eq:label_consistency}
	\min_{\mathbf{\Phi, \mathbf{w}}, \mathbf{\phi}} \sum_{\mathbf{x}_i \in \mathcal{E}}  \ell (\mathbf{w}(\tilde{\mathbf{z}}_i), y_i).
\end{equation}

\subsection{Loss Function}
\label{sec:loss}
Our method is formulated as the following optimization problem:
\begin{equation}
	\label{eq:virm}
	\begin{aligned}
		\mathcal{L}_{\text{VIRM}}(\Phi, \mathbf{w}, \phi)
		 & = \mathcal{L}_{\text{IRM}} + \alpha \mathcal{L}_{\phi}.
	\end{aligned}
\end{equation}

\Cref{eq:virm} consists of two components: the invariant risk and the loss function of the domain-shared SDA module $\mathcal{L}_{\phi}$. The loss function of the domain-shared SDA module is defined as:
\begin{equation}
	\begin{aligned}
		 & \mathcal{L}_{\phi} =
		 & - \frac{1}{2}\sum_{i=0}^n(1 + \log(\boldsymbol{\sigma}^2) - \boldsymbol{\sigma}^2 ) +
		\frac{1}{2n}\sum_{l=1}^n (\mathbf{\hat{z}_i} - \mathbf{z}_i)^2.
		\label{eq:loss_dssda}
	\end{aligned}
\end{equation}

For the first term $\mathcal{L}_{\text{IRM}}$ in \Cref{eq:virm}, there are several methods to implement IRM.
However, our experiments align with the results of the VIRM framework, where VREx~\cite{krueger2021out} was adopted as the specific implementation for IRM.
VREx improves upon previous IRM formulations by introducing a penalty term for the risk variance, thus encouraging the model to reduce risk disparities across domains, and enhancing its robustness to changes in the underlying distribution.
\section{Experiments}
\label{sec:method}

\begin{table*}[htbp]
    \caption{
        Comparison of the proposed method with the state-of-the-art methods on GRDBench~\cite{dgdr2023} dataset.
        Best and second-best results are highlighted in \colorbox[HTML]{ACDBDF}{\textbf{Best}} and \colorbox[HTML]{f0ece2}{\textbf{Second-Best}}, respectively.
    }
    \renewcommand{\arraystretch}{1.2}
    \begin{center}
        \resizebox{\textwidth}{!}{
            \begin{tabular}{l|ccc|ccc|ccc|ccc|ccc}
                \hline
                \multirow{2}{*}{Method}          & \multicolumn{3}{c}{\textbf{DEEPDR}} & \multicolumn{3}{c}{\textbf{APTOS}} & \multicolumn{3}{c}{\textbf{RLDR}} & \multicolumn{3}{c}{\textbf{IDRID}} & \multicolumn{3}{c}{\textbf{Average}}                                                                                                                                                                                                                                                                     \\
                \cline{2-16}
                                                 & AUC                                 & ACC                                & F1                                & AUC                                & ACC                                  & F1                      & AUC                     & ACC                     & F1                      & AUC                     & ACC                     & F1                      & AUC                     & ACC                     & F1                      \\
                \hline
                ERM~\cite{vapnik1999nature}      & $0.676$                             & $0.326$                            & $0.137$                           & $0.640$                            & $0.201$                              & $0.067$                 & $0.755$                 & $0.324$                 & $0.099$                 & $0.649$                 & $0.535$                 & $0.174$                 & $0.680$                 & $0.347$                 & $0.119$                 \\
                CABNet~\cite{he2020cabnet}       & $0.634$                             & $0.417$                            & $0.188$                           & $0.558$                            & $0.202$                              & $0.067$                 & $0.660$                 & $0.324$                 & $0.098$                 & $0.594$                 & $0.468$                 & $0.175$                 & $0.612$                 & $0.353$                 & $0.132$                 \\
                Fishr~\cite{rame2021ishr}        & \cellcolor{1st} $0.794$             & \cellcolor{1st} $0.547$            & \cellcolor{2nd} $0.302$           & $0.774$                            & $0.232$                              & $0.149$                 & $0.804$                 & $0.355$                 & $0.190$                 & $0.726$                 & \cellcolor{1st} $0.517$ & $0.272$                 & $0.774$                 & \cellcolor{1st} $0.413$ & $0.228$                 \\
                GDRNet~\cite{dgdr2023}           & $0.748$                             & $0.447$                            & \cellcolor{1st} $0.323$           & $0.780$                            & \cellcolor{1st} $0.494$              & \cellcolor{1st} $0.366$ & $0.788$                 & $0.316$                 & \cellcolor{2nd} $0.275$ & $0.733$                 & $0.308$                 & $0.307$                 & $0.762$                 & \cellcolor{2nd} $0.391$ & \cellcolor{1st} $0.318$ \\
                GREEN~\cite{liu2020green}        & $0.742$                             & \cellcolor{2nd} $0.494$            & $0.278$                           & $0.704$                            & $0.202$                              & $0.076$                 & $0.779$                 & $0.347$                 & $0.214$                 & $0.666$                 & \cellcolor{2nd} $0.516$ & $0.224$                 & $0.723$                 & $0.390$                 & $0.198$                 \\
                IRM~\cite{arjovsky2019invariant} & $0.767$                             & $0.297$                            & $0.213$                           & $0.763$                            & $0.222$                              & $0.148$                 & $0.795$                 & \cellcolor{1st} $0.388$ & $0.261$                 & \cellcolor{2nd} $0.740$ & $0.513$                 & \cellcolor{2nd} $0.317$ & $0.766$                 & $0.355$                 & $0.235$                 \\
                VREx~\cite{krueger2021out}       & $0.752$                             & $0.262$                            & $0.183$                           & $0.748$                            & $0.217$                              & $0.124$                 & $0.773$                 & $0.351$                 & $0.191$                 & $0.725$                 & $0.472$                 & $0.309$                 & $0.749$                 & $0.325$                 & $0.202$                 \\
                \hline
                VIRM~\cite{zhu2024enlarging}     & \cellcolor{2nd} $0.774$             & $0.356$                            & $0.220$                           & $0.801$                            & $0.240$                              & $0.190$                 & $0.804$                 & $0.378$                 & $0.283$                 & $0.728$                 & $0.448$                 & $0.276$                 & $0.776$                 & $0.356$                 & $0.242$                 \\
                \hline
                \textbf{MedSDA}                  & $0.756$                             & $0.298$                            & $0.190$                           & \cellcolor{2nd} $0.803$            & \cellcolor{2nd} $0.245$              & \cellcolor{2nd} $0.200$ & \cellcolor{1st} $0.817$ & \cellcolor{2nd} $0.386$ & $0.251$                 & $0.731$                 & $0.508$                 & $0.292$                 & \cellcolor{2nd} $0.777$ & $0.359$                 & $0.233$                 \\
                \textbf{Ours}                    & $0.767$                             & $0.338$                            & $0.213$                           & \cellcolor{1st} $0.806$            & $0.236$                              & $0.185$                 & \cellcolor{2nd}$0.811$  & $0.384$                 & \cellcolor{1st} $0.291$ & \cellcolor{1st} $0.746$ & $0.476$                 & \cellcolor{1st} $0.370$ & \cellcolor{1st} $0.782$ & $0.358$                 & \cellcolor{2nd} $0.265$ \\
                \hline
            \end{tabular}
            \label{tab:comparison}}
    \end{center}
\end{table*}

\subsection{Datasets}

\begin{figure}[htbp]
    \centering
    \includegraphics[width=0.45\textwidth]{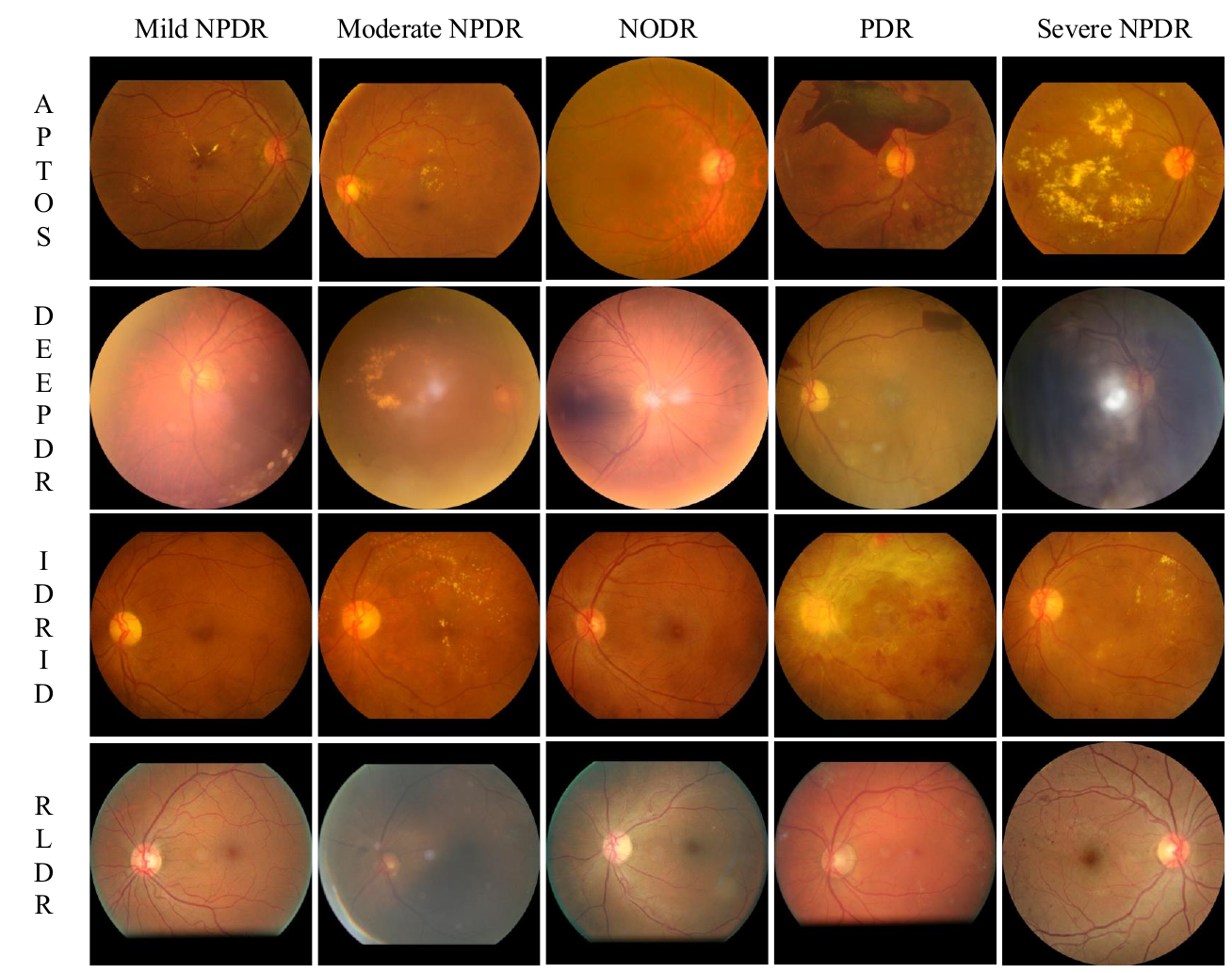}
    \caption{Samples of different center dataset of GRDBench~\cite{dgdr2023}.}
    \label{fig:samples}
\end{figure}

The proposed method is evaluated using the GRDBench dataset \cite{dgdr2023}.
GRDBench, introduced by Che~\etal~\cite{dgdr2023}, serves as a benchmark for domain generalization and consists of eight popular datasets along with two evaluation settings.
Due to redistribution restrictions, our experiments focus on four of these datasets: IDRID~\cite{IDRID}, APTOS~\cite{APTOS}, RLDR~\cite{RLDR}, and DEEPDR~\cite{DEEPDR}.
The datasets used are APTOS (3662 samples), DEEPDR (1999 samples), IDRID (516 samples), and RLDR (1593 samples).
\Cref{fig:samples} shows samples from each dataset, highlighting the diversity and complexity of the images.
Although our method was specifically evaluated on these datasets, its principles of semantic data augmentation and invariant risk minimization suggest potential applicability to other medical image analysis tasks.

\subsection{Evaluation Protocol}

To assess the generalization performance of different methods, we follow the leave-one-domain-out cross-validation strategy recommended by GRDBench~\cite{dgdr2023}.
Specifically, in each experimental run, we designate one of the domains (APTOS, DEEPDR, IDRID, or RLDR) as the test domain, using the remaining three domains for training and validation.
This ensures that the model is evaluated on a domain it has never seen during training, simulating real-world domain shifts and enabling robust comparison.

\subsection{Implementation Details}

For the hyperparameters, the learning rate is set to 0.0001 and the batch size to 256 for all experiments, except for GDRNet~\cite{dgdr2023}, where the learning rate of 0.001 is used.
The best learning rate and batch size were selected from \{ 0.001, 0.0001, 0.0005, 0.00005, 0.00001\} and \{16, 32, 64, 128, 256\}, respectively, based on the validation set.
The Adam optimizer~\cite{kingma2014adam} is used with a weight decay of 0.00001.
All experiments were run for 100 epochs, with the learning rate decayed by a factor of 0.1 at epochs 50 and 75.
The backbone network used for all experiments is ResNet-18~\cite{he2016deep}, with an input size of $224 \times 224$ for the GRDBench dataset.
PyTorch is used for the deep learning framework, and all experiments were conducted on an NVIDIA RTX 3090 GPU.

\subsection{Experiment on GRDBench}
\label{sec:grdbench}

We conducted an extensive evaluation to compare the performance of our proposed method with several state-of-the-art approaches on the GRDBench dataset~\cite{dgdr2023}.
The comparison includes a vanilla baseline method (ERM)\cite{vapnik1999nature}, along with advanced techniques from various domains such as ophthalmic disease diagnosis (Fishr, GDRNet)\cite{rame2021ishr, krueger2021out},
domain generalization (DGT) methods like IRM~\cite{arjovsky2019invariant} and VREx~\cite{krueger2021out}, and feature representation learning methods (GREEN, CABNet)\cite{liu2020green, he2020cabnet}.
These methods were selected based on their relevance to domain generalization and their adaptability to medical image tasks, with minimal to no modifications required for application to our dataset.
For all methods, we used the standard domain generalization augmentation pipeline\cite{dgdr2023} except DRGen, which employed a default augmentation strategy~\cite{dgdr2023}.

\begin{table}[t]
    \caption{Ablation study of the proposed method.
        Best and second-best results are highlighted in \colorbox[HTML]{ACDBDF}{\textbf{Best}} and \colorbox[HTML]{f0ece2}{\textbf{Second-Best}}, respectively.
    }
    \renewcommand{\arraystretch}{1.2}
    \begin{center}
        \begin{tabular}{c|cc|ccc}
            \hline
                          & \multicolumn{2}{c|}{\textbf{Director}} & \multicolumn{3}{c}{\textbf{Average}}                                                                                 \\
            \cline{2-6}
                          & \textbf{MMD~\cite{li2018domain}}       & \textbf{Cov~\cite{wang2024inter}}    & \textbf{\textit{AUC}}   & \textbf{\textit{ACC}}    & {\textbf{\textit{F1}}}   \\
            \hline
            \textbf{Soft} & \checkmark                             &                                      & $0.769$                 & $0.311$                  & $0.218$                  \\
            \textbf{Soft} &                                        & \checkmark                           & $0.781$                 & \cellcolor{1st}  $0.364$ & $0.240$                  \\
            \hline
            \textbf{Hard} & \checkmark                             &                                      & $0.767$                 & $0.341$                  & \cellcolor{2nd}  $0.245$ \\
            \textbf{Hard} &                                        & \checkmark                           & \cellcolor{1st} $0.782$ & \cellcolor{2nd}  $0.358$ & \cellcolor{1st} $0.265$  \\
            \hline
        \end{tabular}
        \label{tab:ablation}
    \end{center}
\end{table}

\textbf{Results:}
The results, shown in \Cref{tab:comparison}, demonstrate that our method outperforms all other approaches across multiple metrics (AUC, ACC, F1) on most test domains, particularly on DEEPDR, APTOS, and RLDR.
Notably, our method achieved the highest AUC in these domains, surpassing other methods by a significant margin.
While some methods like Fishr~\cite{rame2021ishr} and GDRNet~\cite{dgdr2023} performed well in certain tests, our method consistently outperformed them in key areas, such as the average AUC across the four test domains.
The results also reveal that domain generalization methods (such as IRM~\cite{arjovsky2019invariant} and VREx~\cite{krueger2021out}) show significant improvements over the baseline ERM, but our approach further refines generalization by learning more robust features that preserve diagnostic patterns while increasing intra-class variation.
Furthermore, compared to the feature representation learning methods like GREEN~\cite{liu2020green} and CABNet~\cite{he2020cabnet}, our method shows notable improvements in both accuracy and robustness, making it a superior choice for domain generalization in medical image tasks.
These results underscore the effectiveness of our approach in addressing domain shifts and improving generalization performance across multiple medical imaging datasets.

\begin{figure}[t]
    \centering
    \includegraphics[width=0.45\textwidth]{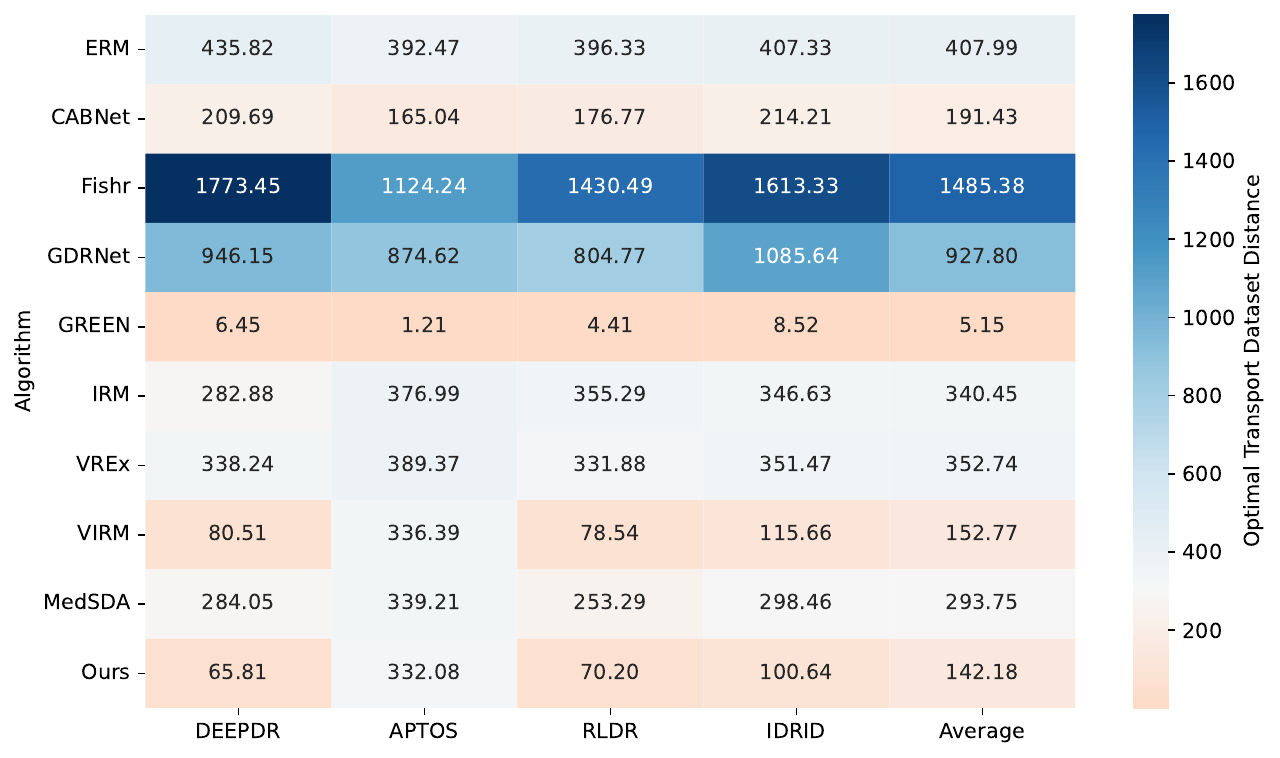}
    \caption{
        Heatmap of the Optimal Transport Dataset Distance (OTDD)\cite{alvarez2020geometric} for different domain pairs in the GRDBench dataset~\cite{dgdr2023}.
    }
    \label{fig:ootd_heatmap}
\end{figure}

\subsection{Ablation Study}
\label{sec:ablation}

\begin{figure*}[t]
    \centering
    \begin{subfigure}[b]{0.45\textwidth}
        \centering
        \includegraphics[width=\textwidth]{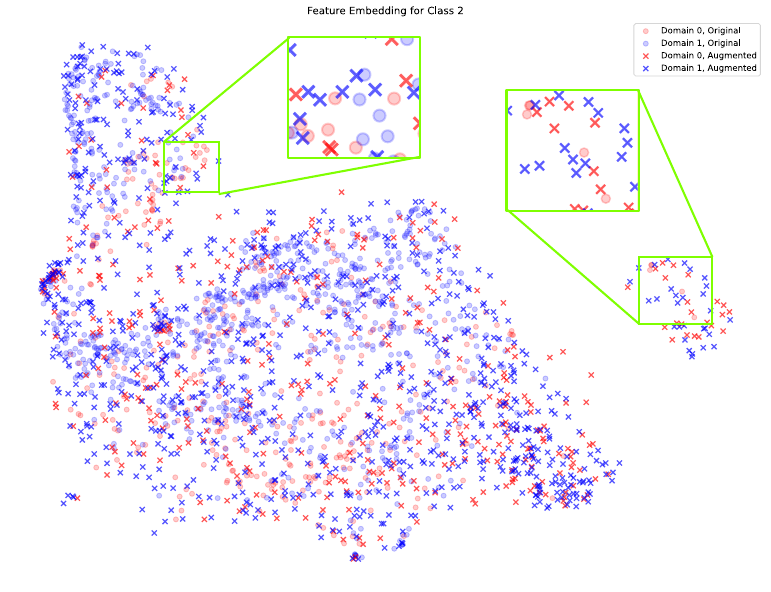}
        \caption{VIRM}
        \label{fig:vis_aug_dir_virm}
    \end{subfigure}
    \hfill
    \begin{subfigure}[b]{0.45\textwidth}
        \centering
        \includegraphics[width=\textwidth]{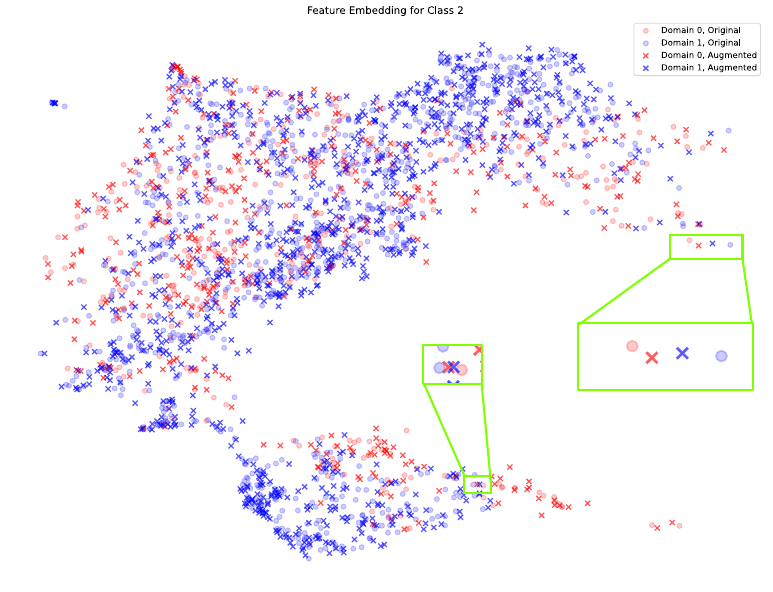}
        \caption{Ours}
        \label{fig:vis_aug_dir_ours}
    \end{subfigure}
    \caption{
        Visualizing deep features on GRDBench dataset with two domains using UMAP\cite{mcinnes2018umap}.
        For easier demonstration of the effectiveness of the proposed method, we visualized data from only two domains, although the original setting included four domains.
        As shown in the green box in \Cref{fig:vis_aug_dir_ours}, the proposed method consistently enhances features towards the target domain.
        In contrast, the VIRM method (as shown in \Cref{fig:vis_aug_dir_virm}) produces many random and meaningless directions, especially in the small-batch data scenario, leading to less stable feature representations.
    }
    \label{fig:vis_aug_direction}
\end{figure*}

\begin{figure*}[htbp]
    \centering
    \begin{subfigure}[b]{0.19\textwidth}
        \centering
        \includegraphics[width=\textwidth]{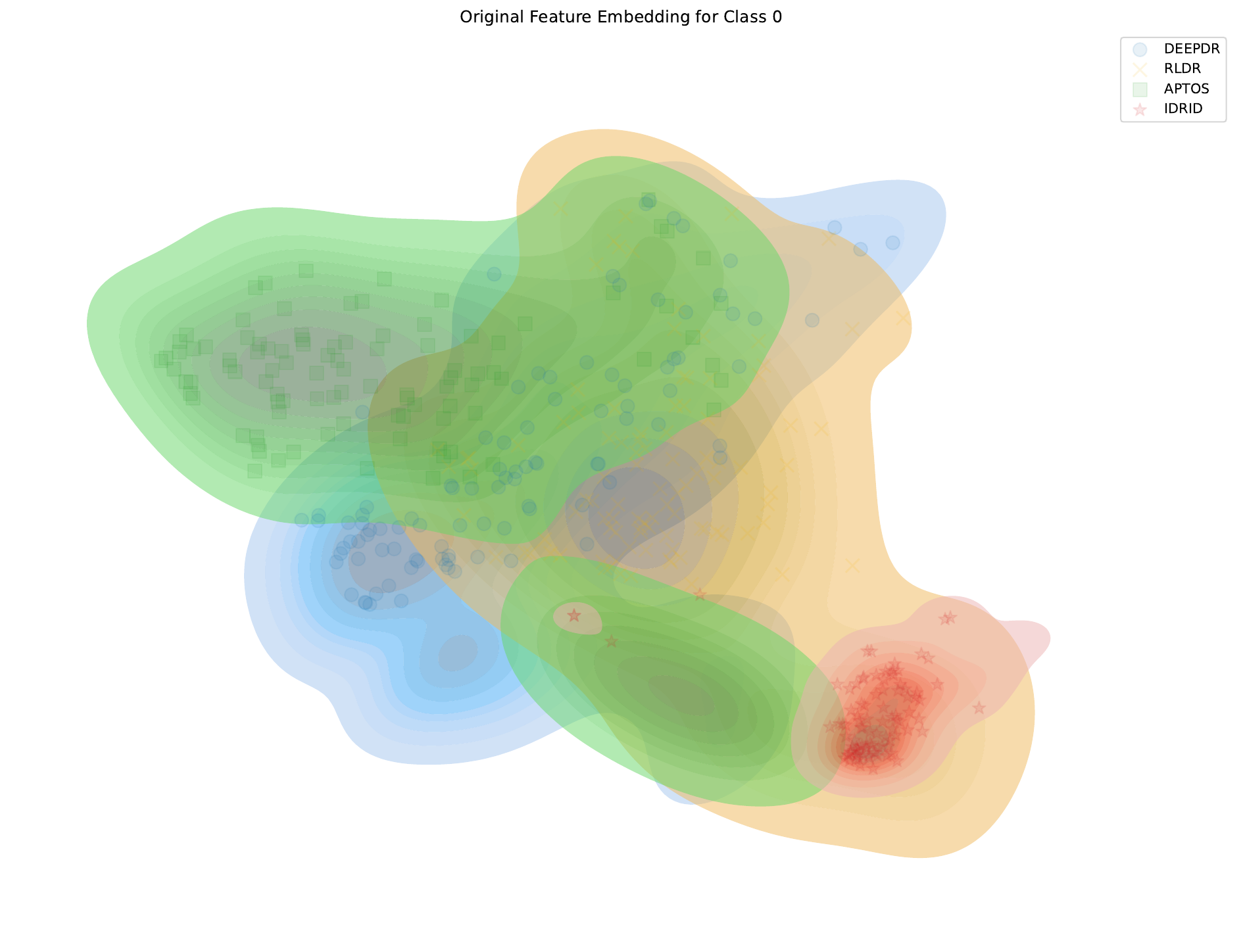}
        \caption{Class 0}
        \label{fig:vis_kde_0}
    \end{subfigure}
    \hfill
    \begin{subfigure}[b]{0.19\textwidth}
        \centering
        \includegraphics[width=\textwidth]{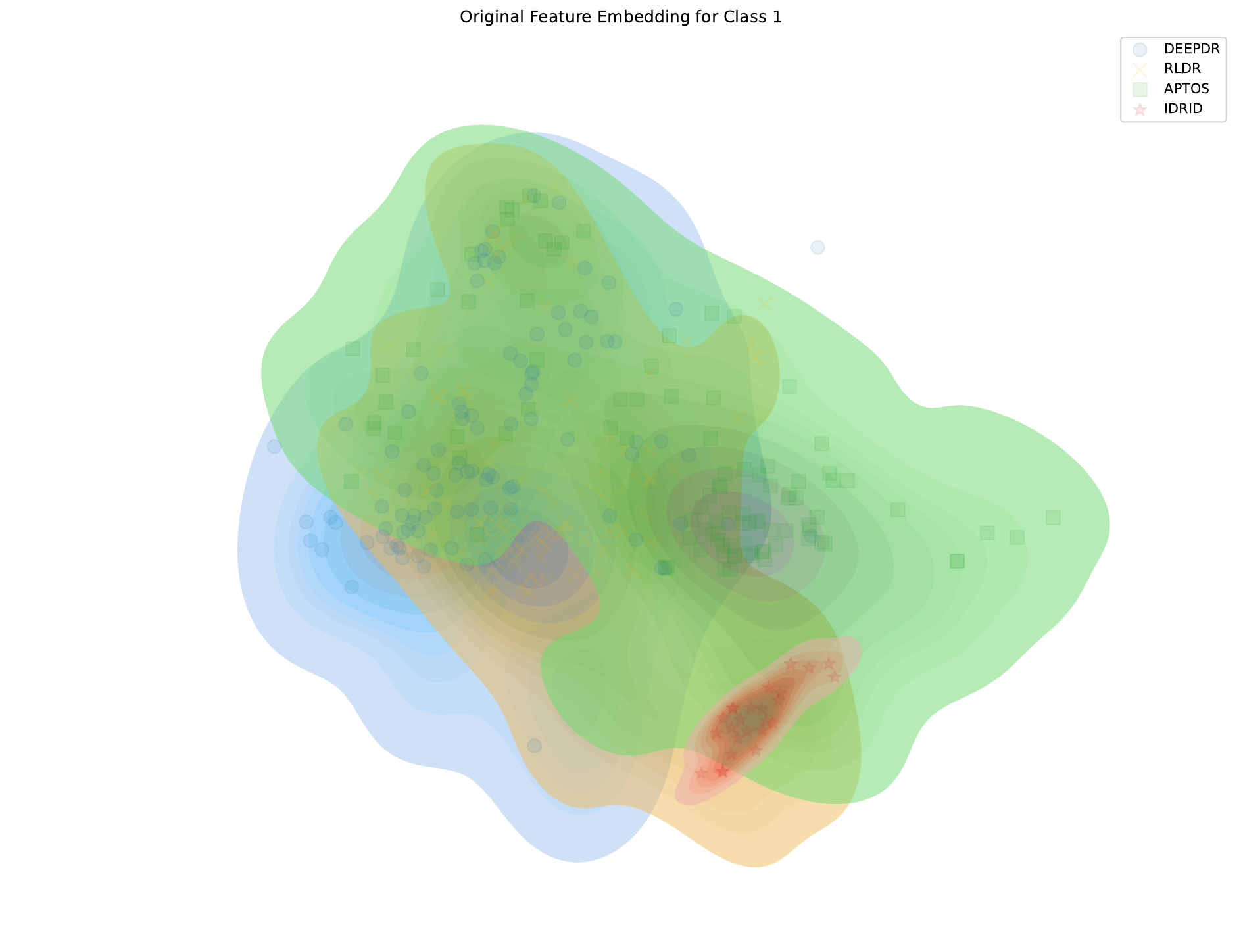}
        \caption{Class 1}
        \label{fig:vis_kde_1}
    \end{subfigure}
    \hfill
    \begin{subfigure}[b]{0.19\textwidth}
        \centering
        \includegraphics[width=\textwidth]{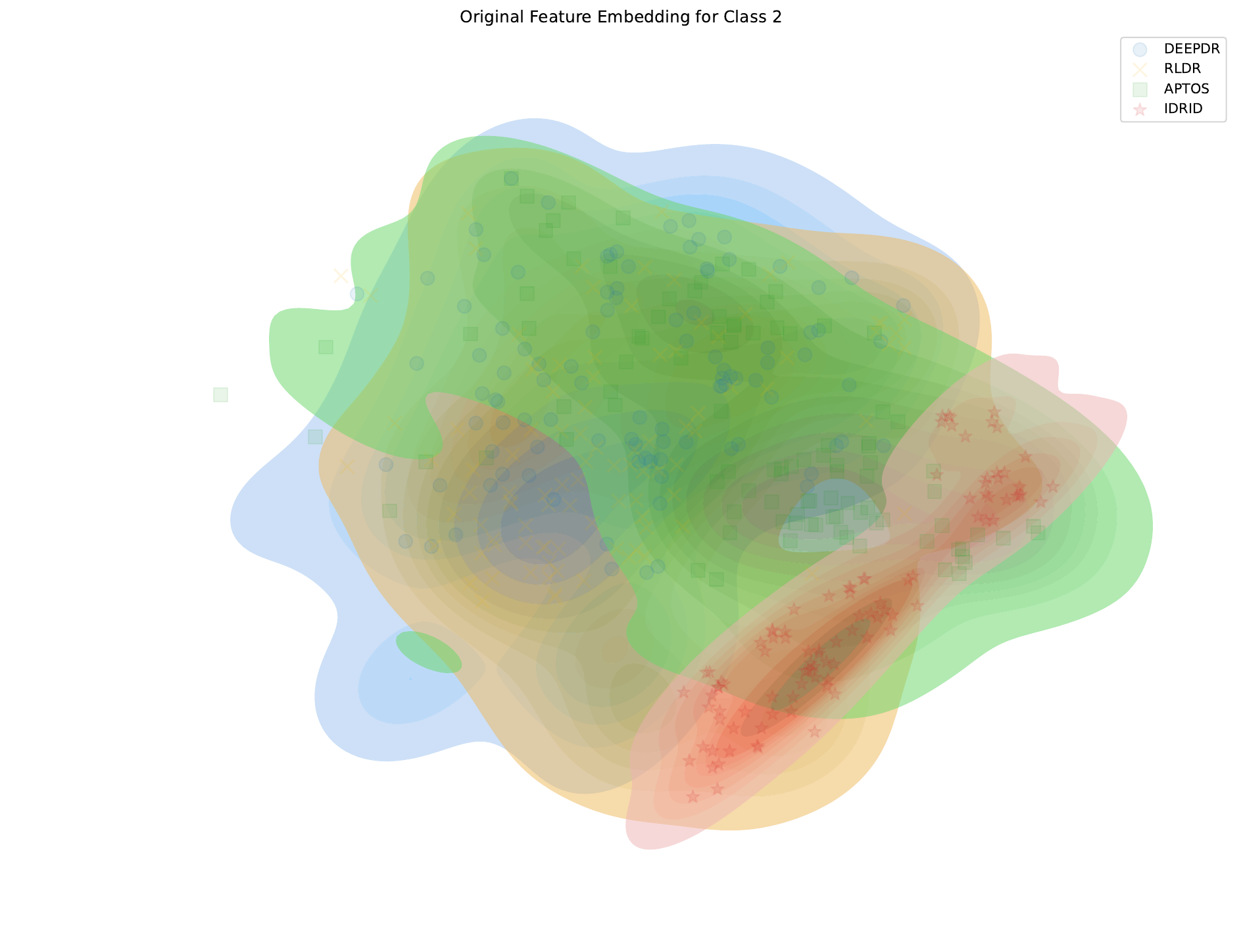}
        \caption{Class 2}
        \label{fig:vis_kde_2}
    \end{subfigure}
    \hfill
    \begin{subfigure}[b]{0.19\textwidth}
        \centering
        \includegraphics[width=\textwidth]{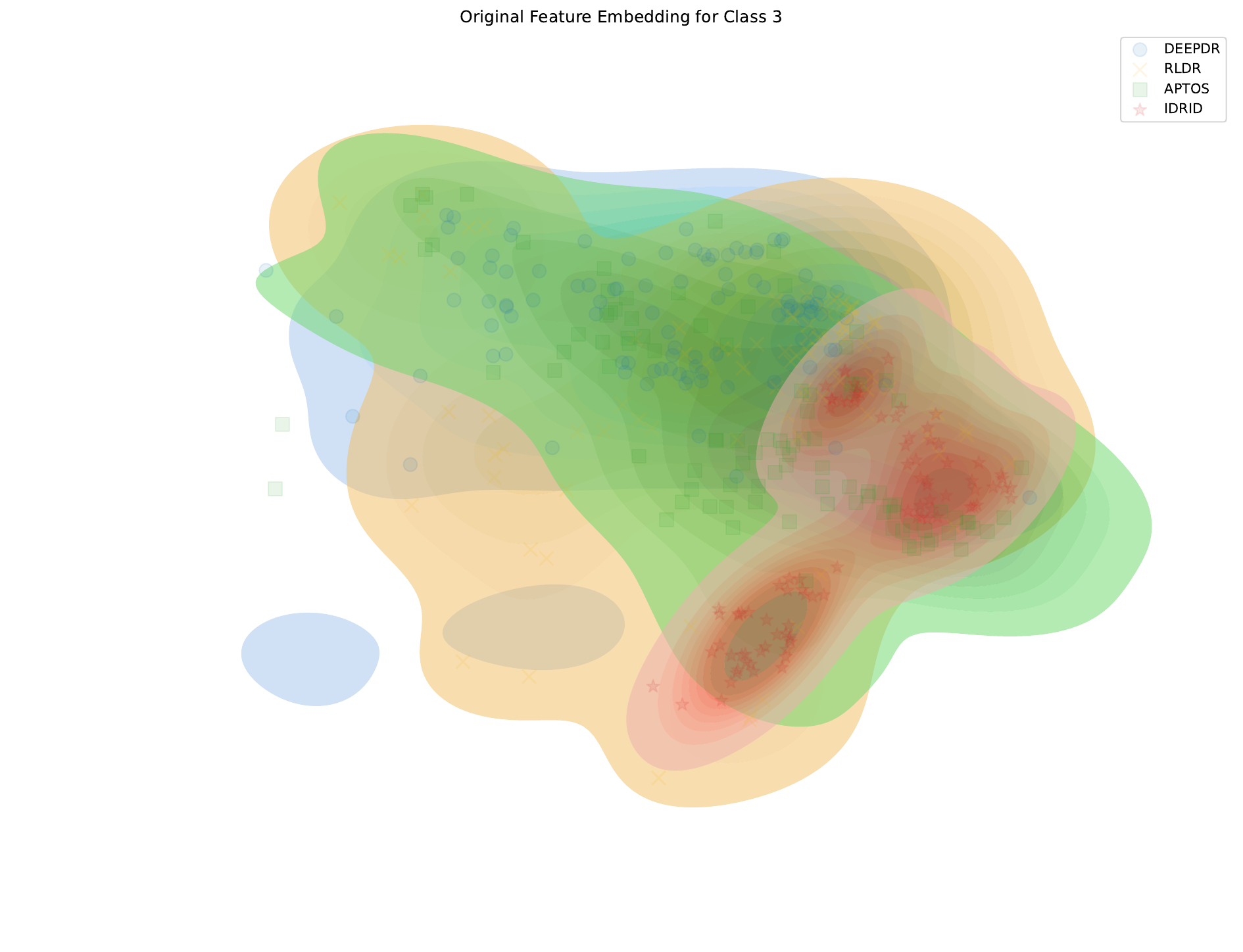}
        \caption{Class 3}
        \label{fig:vis_kde_3}
    \end{subfigure}
    \hfill
    \begin{subfigure}[b]{0.19\textwidth}
        \centering
        \includegraphics[width=\textwidth]{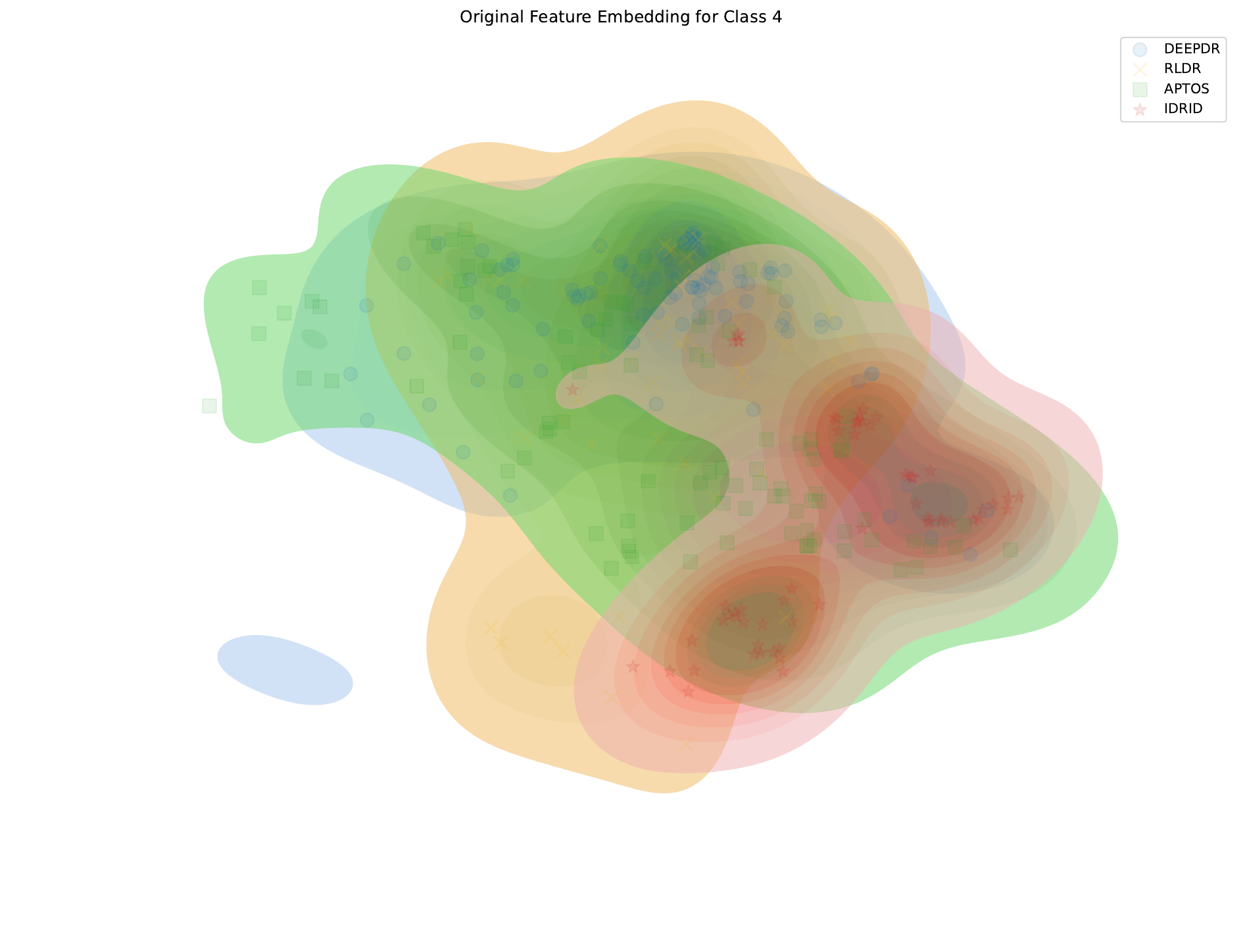}
        \caption{Class 4}
        \label{fig:vis_kde_4}
    \end{subfigure}
    \caption{
        KDE-Based Visualization of Deep Features Using UMAP\cite{mcinnes2018umap}.
    }
    \label{fig:vis_kde}
\end{figure*}

The goal of the direction selector is to make domain A move in the direction of domain B.
This can be achieved by optimizing the direction selector by calculating the difference between the augmented features and other domain features.
We consider two potential methods for direction selectors, including maximum mean difference (MMD~\cite{li2018domain}), covariance of the different source domains (Cov~\cite{wang2024inter}), and domain discriminator~\cite{bai2021decaug}.
Additionally, we consider potential two different settings for the direction selector: \textbf{Soft} and \textbf{Hard}.
\textbf{Soft} indicates that $\mathbb{d} \in [0, 1]$ and \textbf{Hard} indicates that $\mathbb{d} \in \{0, 1\}$.
\Cref{tab:ablation} shows the results of the ablation study, where we compare the performance of our method with different direction selectors and settings.

\subsection{Analysis via OTDD}
\label{sec:ootd}

To further evaluate the proposed method’s ability to reduce domain discrepancies, we employed the Optimal Transport Dataset Distance (OTDD)\cite{alvarez2020geometric}.
OTDD measures the distance between two datasets using optimal transport, where smaller values indicate reduced domain differences.
Alvarez\etal~\cite{alvarez2020geometric} demonstrated a linear relationship between OTDD and transferability performance, showing a strong and significant correlation, which highlights OTDD’s predictive power for transferability across datasets.

We innovatively apply OTDD to assess the generalization between two distinct domains within the same dataset.
As shown in \Cref{fig:ootd_heatmap}, our method, excluding the GREEN method~\cite{liu2020green}, achieves the lowest average OTDD across domain pairs, indicating superior generalization across diverse domains.
The GREEN method~\cite{liu2020green} performs better in OTDD due to its use of graph convolutional networks (GCNs) to model inter-class dependencies.
Since OTDD evaluates label-to-label distances, this results in improved OTDD performance for GREEN.
However, our method exhibits significantly better classification performance compared to GREEN, highlighting its practicality in classification tasks.

Another key observation is in the last two rows, emphasizing the importance of incorporating invariant risk constraints.
This also indirectly supports the perspective presented in VIRM.
Furthermore, when compared to VIRM, our approach achieves better OTDD results, reinforcing its enhanced generalization capability across domain pairs.

\subsection{Visualization}
\label{sec:visualization}

\subsubsection{Compare with VIRM}

To further emphasize the core advantages of our method, we compared its feature space visualization with that of the VIRM\cite{zhu2024enlarging} method. As shown in the green box in \Cref{fig:vis_aug_dir_virm}, VIRM employs a strategy of randomly selecting augmentation directions, which results in a more dispersed feature space visualization.
This suggests that the feature representations learned by VIRM are less stable and compact in the feature space. While this approach may suffice for large datasets that can cover most of the feature space, it becomes less robust for small-batch datasets.

In small-batch settings, the randomness in selecting augmentation directions can lead to meaningless transformations, exacerbating the dispersion of feature space representations. In contrast, as illustrated in the green box in \Cref{fig:vis_aug_dir_ours}, our method introduces a domain-covariance-guided augmentation direction selector.
This ensures that features are consistently enhanced toward the target domain, resulting in a tighter and more stable feature space overlap.
As a result, the feature space visualization of our method is more compact and stable.

Zhu \etal~\cite{zhu2024enlarging} proposed the use of semantic data augmentation to improve IRM-based methods by enhancing feature overlap through semantic diversity, thereby improving model generalization in large-scale datasets.
Similarly, our method demonstrates superior performance on small-batch medical datasets by guiding feature enhancement toward the target domain. This ensures more stable and compact feature space representations, which facilitate better generalization.

While \Cref{fig:meaningless} provides an idealized model, \Cref{fig:vis_aug_direction} presents the visualization of our method applied to real-world datasets.
These results further validate the effectiveness of our approach and, more specifically, the augmentation direction selector we proposed.

\subsubsection{KDE Visualization}

Feature support overlap is a crucial determinant of successful generalization\cite{ahuja2021invariance}.
Furthermore, it reflects the extent of dataset diversity shifts along the feature dimensions.
To evaluate this overlap, we employed kernel density estimation (KDE)\cite{wkeglarczyk2018kernel}, which provides a clear visualization of the feature space overlap.
As shown in \Cref{fig:vis_kde}, our method demonstrates notable overlap in the GRDBench dataset, offering an intuitive representation of its effectiveness in reducing domain differences and enhancing generalization.

\subsubsection{Feature Visualization}

\begin{figure}[htbp]
    \centering
    \includegraphics[width=0.45\textwidth]{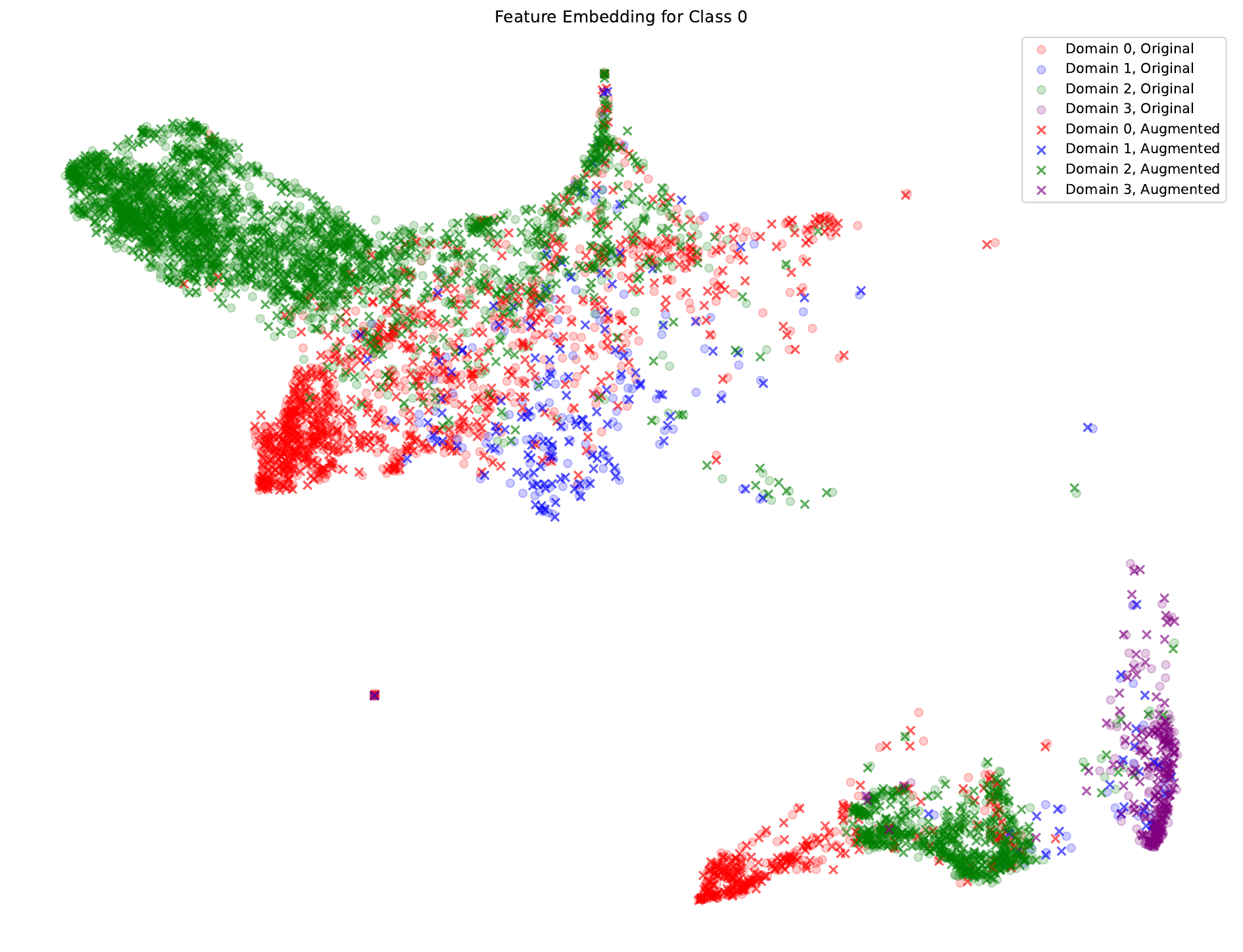}
    \caption{
        UMAP\cite{mcinnes2018umap}. Visualization of Class 0 Deep Features on the GRDBench Dataset\cite{dgdr2023} (Original Features Only).
    }
    \label{fig:vis_class_0}
\end{figure}

We also visualized the distribution of features from all domains within the same class in the feature space.
As illustrated in \Cref{fig:vis_class_0}, the features enhanced by our method consistently move toward directions that align with other domains, rather than following random directions.
This alignment results in a more compact feature space overlap, which significantly contributes to improving the generalization performance of the model.

\section{Conclusion}
\label{sec:conclusion}

In this work, we proposed a novel domain-oriented framework to enhance the generalization performance of invariant risk minimization (IRM) by expanding the feature overlap between domains.
Specifically, we introduced an augmentation direction selector that enhances feature diversity while maintaining label consistency.
Our method was not specifically designed for the diabetic retinopathy grading task, but rather as a general framework for domain generalization in medical image analysis.
This flexibility suggests that our approach has the potential for broad applicability across different medical imaging domains.
Our experiments on challenging datasets demonstrated significant improvements in accuracy and robustness compared to baseline methods, highlighting the potential of our approach for real-world domain generalization tasks.
Despite promising results, the method was primarily evaluated on synthetic and small-scale datasets, and its scalability in large-scale scenarios needs further investigation.
Future work should explore more efficient implementations and evaluate the framework on diverse, real-world medical datasets.

\bibliographystyle{IEEEtran}
\bibliography{ref}

\end{document}